\pdfoutput=1
\documentclass[letterpaper, 10 pt, conference]{ieeeconf}  
\IEEEoverridecommandlockouts                              
\usepackage{amsmath,amsfonts}
\usepackage{algorithmic}
\usepackage{algorithm}
\usepackage{array}
\usepackage[caption=false,font=normalsize,labelfont=sf,textfont=sf]{subfig}
\usepackage{textcomp}
\usepackage{stfloats}
\usepackage{url}
\usepackage{verbatim}
\usepackage{graphicx}
\usepackage{cite}
\usepackage{colortbl}  
\usepackage{float}
\usepackage{booktabs}
\usepackage{bbding}
\usepackage{multirow}
\usepackage{eccvabbrv}
\usepackage{cite}
\usepackage{hyperref}

\newcommand{\zdx}[1]{{#1}}
\newcommand{\re}[2]{{#2}}

\title{\LARGE \bf
Advancing Object-Goal Navigation through LLM-enhanced Object Affinities Transfer
}

\author{
    Mengying Lin$^{1, 2, \dagger}$, Shugao Liu$^{1, \dagger}$, Dingxi Zhang$^{3}$, Yaran Chen$^{1, 4}$, Zhaoran Wang$^{5}$, Haoran Li$^{1, *}$, Dongbin Zhao$^{1}$ \\
    \thanks{$^{1}$Institute of Automation, Chinese Academy of Sciences, Beijing, China.}
    \thanks{$^{2}$Georgia Institute of Technology, Atlanta, US.}
    \thanks{$^{3}$ETH Zurich, Zurich, CH.}
    \thanks{$^{4}$Xi'an Jiaotong-Liverpool University, Suzhou, China.}
    \thanks{$^{5}$McCormick School of Engineering,  Northwestern University.}
    \thanks{${\dagger}$Authors contribute equally to this work.}
    \thanks{*Corresponding author {\tt\small lihaoran2015@ia.ac.cn}}
    \thanks{This work is partly supported by the National Natural Science Foundationof China (NSFC) under Grants No. 62173324, and in part by the CAS for Grand Challenges under Grant 104GJHZ2022013GC.}
}

\begin{document}
\maketitle
\thispagestyle{empty}
\pagestyle{empty}

\begin{abstract}
Object-goal navigation requires mobile robots to efficiently locate targets with visual and spatial information, yet existing methods struggle with generalization in unseen environments. Heuristic approaches with naive metrics fail in complex layouts, while graph-based and learning-based methods suffer from environmental biases and limited generalization.
Although Large Language Models (LLMs) as planners or agents offer a rich knowledge base, they are cost-inefficient and lack targeted historical experience. To address these challenges, we propose the LLM-enhanced Object Affinities Transfer (LOAT) framework, integrating LLM-derived semantics with learning-based approaches to leverage experiential object affinities for better generalization in unseen settings. LOAT employs a dual-module strategy: one module accesses LLMs' vast knowledge, and the other applies learned object semantic relationships, dynamically fusing these sources based on context. Evaluations in AI2-THOR and Habitat simulators show significant improvements in navigation success and efficiency, and real-world deployment demonstrates the zero-shot ability of LOAT to enhance object-goal navigation systems.

\end{abstract}
\section{Introduction}
Object navigation is a basic skill for the mobile robot, which enables the robot to navigate to a specified target object within a scene given category names. The objective is to find the object successfully and minimize the navigation path \cite{1sun2024survey}. Considering the trade-off between success rate and efficiency, how to leverage object semantic relationships in the household environment is an important direction for more efficient navigation.


Understanding object relationships is key to efficient navigation. Heuristic methods using non-training distance metrics like Euclidean distance \cite{chen2023not-train} are simple but fail to reflect navigable path lengths in complex multi-room layouts. Graph-based methods construct learnable nodes and edges representing categories and correlations \cite{zhang2021hierarchical}, yet struggle with generalization due to training biases and stationary structures that cannot handle unseen targets.


Learning-based methods using elaborate training data poorly generalize to new scenes or objects. Large language models (LLMs) and vision-language models (VLMs) offer richer contextual insights to enhance generalization \cite{zhou2023esc}, but inconsistent inference scores and variable object priorities with paraphrased prompts affect stability. Direct LLM/VLM applications as navigation agents (SayNav \cite{rajvanshi2023saynav}, NavGPT \cite{zhou2023navgpt}) show strong generalization but require frequent, computationally intensive queries. While excelling in general reasoning, their broad knowledge may lack specificity for certain contexts like culturally specific household layouts, potentially causing suboptimal performance.




In this work, we propose the LLM-enhanced Object Affinities Transfer (LOAT) framework to improve generalization and navigation performance in novel environments. LOAT integrates LLM commonsense reasoning through a generalized affinities module with learned object relationships from an experiential affinities module. This combination leverages LLMs' diverse cultural knowledge while grounding it with experiential patterns, enhancing decision-making in complex household layouts. A dynamic fusion module balances these affinity sources based on temporal contexts.


Our main contributions are:
\begin{itemize}
    \item We propose LOAT, a novel framework that enhances object-goal navigation efficiency by providing object affinity guidance to downstream navigation policies with minimal architectural modifications.
    \item We design three core modules—experiential affinities, generalized affinities, and dynamic fusion—that significantly improve generalization for unseen objects through effective LLM-historical information integration.
    \item Extensive experiments across multiple simulators and real-world tasks demonstrate LOAT substantially improves navigation success rates and path efficiency with impressive generalization in novel scenarios.
\end{itemize}


\section{Related Work}
\subsection{Traditional Methods to Object-Goal Navigation}

Object-goal navigation, a foundational area in Embodied AI, aims to enable agents to locate specific target objects in unseen environments. Traditional approaches fall into two categories: end-to-end learning-based methods that train agents directly through sensory inputs, and map-based methods that construct environmental representations to guide navigation.

\subsubsection{End-to-End Learning-Based Methods}

End-to-end learning methods have gained traction for training embodied navigation agents~\cite{mat}, focusing on maximizing reward functions that incentivize successful object discovery~\cite{Haoran2020}.Recent advancements introduce novel architectures enhancing navigation capabilities. Decision Transformer~\cite{chen2021decisiontransformer} employs sequence modeling to learn policies by conditioning on historical data, enabling adaptability across navigation tasks. MTVM ~\cite{lin2021multimodal} explicitly models historical observation-instruction interactions, crucial for understanding navigation trajectory progress. PoliFormer~\cite{zeng2024poliformer} represents a fully transformer-based agent integrating causal transformer decoders with long-term memory and reasoning capabilities, designed to scale with on-policy reinforcement learning training.

\subsubsection{Map-Based Methods}
Map-based methods fall into two categories: direct prediction and frontier-based exploration.
\paragraph{Direct Prediction}

Direct prediction includes action prediction and waypoint prediction. Action prediction uses current observations and semantic maps to guide real-time agent movements, enabling responsive adaptation to environmental changes \cite{chen2022duet}. Waypoint prediction trains models on map datasets to generate probabilistic maps indicating potential target locations, facilitating efficient path planning by directing agents toward high-probability areas \cite{min2021film, zhai2023peanut, chen2024common}.

\paragraph{Frontier-based Exploration}
This approach utilizes semantic data to define exploration boundaries through obstacle maps constructed from egocentric RGB-D images \cite{Yamauchi_2002}. By establishing a structured framework, frontier-based methods enhance agents' navigation capabilities, facilitating targeted exploration and improving overall navigation efficiency \cite{gadre2022cow, Ramakrishnan22}.

\subsection{Large Models-based Object Navigation}
Incorporating large models into navigation systems follows three methodologies:

\subsubsection{Object Relevance Scoring with Scripted Strategies} 

This approach leverages large models to evaluate object relevance using scripted search strategies like frontier-based methods\cite{topiwala2018frontier, yuanwen2025}. ESC\cite{zhou2023esc} employs LLMs for direct candidate scoring, while L3MVN\cite{Yu_l3mvn2023} and Prompter\cite{inoue2022prompter} derive collocation probabilities from masked language models with prompts like “Something you find at \texttt{[MASK]} is \texttt{[TARGET]},” scoring candidates based on their placement in the "\texttt{[MASK]}" token. However, model sensitivity to language changes causes significant score variability with prompt modifications, affecting consistency of object relevance assessments.
    
\subsubsection{High-Level Planning with Large Models} 

This method uses large models for strategic planning and task decomposition in long-horizon navigation~\cite{robogpt}. LLM-Planner \cite{song2023llmplanner} breaks down tasks and identifies key landmarks from instructions, establishing subgoals and initiating replanning upon failure. This strategy struggles in scenarios lacking explicit high-level instructions for targeting objects.
    
\subsubsection{Navigation Agent} 

SayNav\cite{rajvanshi2023saynav} and NavGPT\cite{zhou2023navgpt} directly use language models as navigational agents, requiring frequent prompts at every step. While conceptually appealing for dynamic adaptability, computational and time costs constrain practicality for wide-scale application.

\begin{figure*}
  \centering
  \includegraphics[width=1\textwidth]{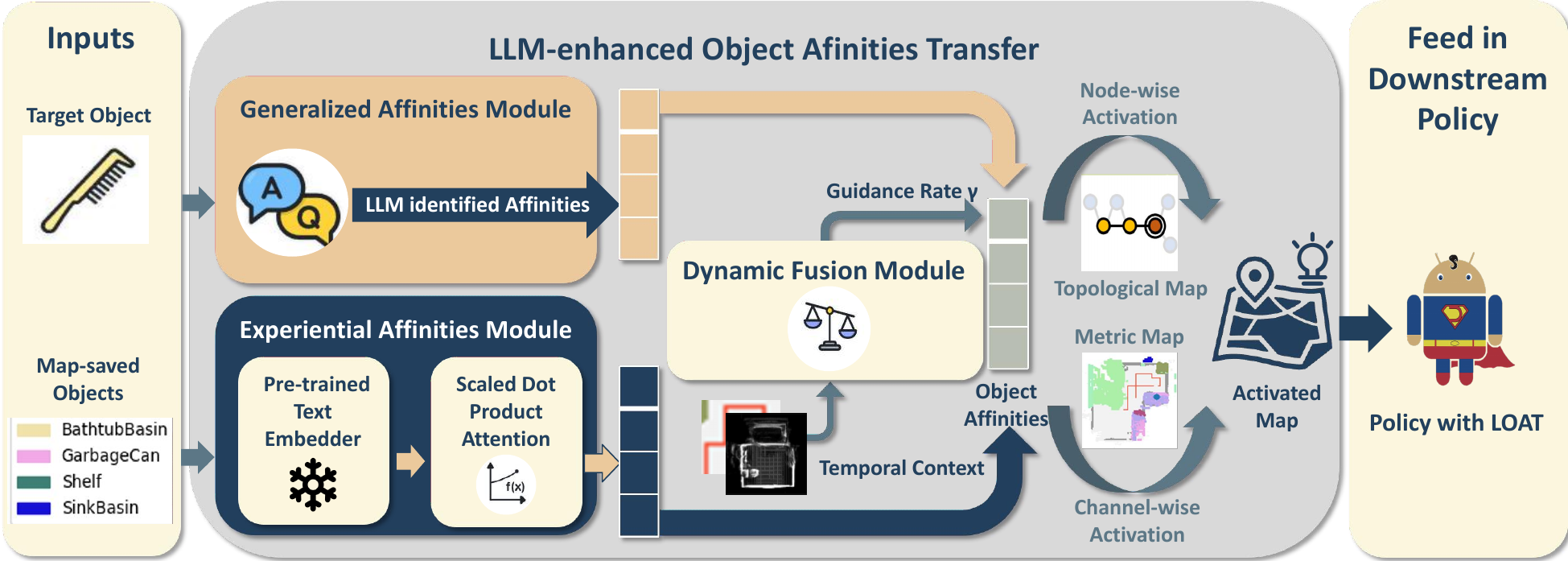} 
  \caption{\textbf{LOAT Framework.} This framework processes the target and scene map categories to locate the target based on object affinities. Category information is passed to two main modules: the generalized affinities module, powered by LLMs, which assesses each object's relevance, and the experiential affinities module, which uses a pre-trained text tokenizer for historical object affinities. A dynamic fusion module balances these inputs using temporal state embeddings before applying the scores for node-wise or channel-wise activation to enhance downstream policy decision-making.}
  \label{fig:loat-detail-arc}
  \vspace{-0.5cm}
\end{figure*}

\section{Methods}

Rather than utilizing LLMs as high-level planners requiring complex alignment with low-level policies, our approach employs language-based priors in numerically interpretable formats for downstream models. LOAT converts semantic object relationships into continuous affinity scores directly integrated into existing navigation architectures. As shown in Fig. \ref{fig:loat-detail-arc}, the LOAT framework integrates LLM-derived insights with historical object affinity data to enhance semantic maps, bridging high-level semantic understanding with low-level policy execution.


The framework consists of three modules: the generalized affinities module captures broad semantic knowledge from LLMs and outputs relevance scores for each object category; the experiential affinities module leverages learned object relationships from training data; and the dynamic fusion module balances these sources based on contextual inputs. These numerical affinity scores directly inform node-wise or channel-wise activation based on map types, enriching semantic maps for informed navigation without requiring architectural changes to underlying policy networks.

\subsection{Experiential Affinities Module}

This module aims at extracting object affinities from training time. It employs a scaled dot-product similarity mechanism to determine the relevance of each environmental object to the target, leveraging pre-trained text embeddings. The set of objects in maps is represented as $\mathcal{O} = \{o_1, o_2,..., o_M\}$, where $M$ is the total category number in maps and $o_{target}$ is the target. The embedding for any object $o_i$ is obtained through a pre-trained text encoder as $\mathbf{e}(o_i)$, which converts object category names into fixed-dimensional vector representations.


To compute the affinity scores, we first transform the embeddings into queries $\mathbf{Q}$ and keys $\mathbf{K}_i$ using learned linear transformations for the target and all other objects respectively:
\begin{equation}
\mathbf{Q} = \mathbf{W}_q\mathbf{e}(o_{target}), \quad \mathbf{K}_i = \mathbf{W}_k\mathbf{e}(o_i),
\end{equation}
where $\mathbf{W}_q$ and $\mathbf{W}_k$ are weight matrices for queries and keys, and $\mathbf{K}_i$ represents the key vector for object $o_i$.


The experiential affinity score $A_{E_i}$ for each object $o_i$ is then calculated using the scaled dot-product similarity with softmax normalization:
\begin{equation}
A_{E_i} = \frac{\exp(\mathbf{Q} \cdot \mathbf{K}_i^T / \sqrt{d_k})}{\sum_{j=1}^{M} \exp(\mathbf{Q} \cdot \mathbf{K}_j^T / \sqrt{d_k})},
\label{eq:2}
\end{equation}
where $d_k$ is the dimensionality of the key vectors, serving to scale the dot product such that it leads to more stable gradients, with $\mathbf{K}_j$ being the key corresponding to the $j^{th}$ object in the environment. This formulation computes continuous similarity scores based on learnable transformations of pre-trained embeddings, capturing statistical patterns observed during training.


This mechanism effectively captures the semantic relationship between the target and every other object in the map by computing normalized similarity scores that prioritize objects based on their learned relevance to the target. The resultant affinity scores guide the module to emphasize features from objects more closely related to the target in training experiences, thus enhancing pattern recognition and facilitating more informed navigation decisions within known contexts.

\subsection{Generalized Affinities Module}

The generalized affinities module leverages semantic relations derived from LLMs to enhance focus on objects semantically related to a specified target. While LLMs could score object affinities directly, inconsistent inference scores and varied object priorities from prompting present challenges. Paraphrased prompts lead to unstable scoring, potentially undermining system reliability. To address this, we shift focus from specific scores to object relevance, ensuring consistent generalized knowledge. Therefore, we use LLM priors solely as identifiers of relevant objects rather than for exact scoring.



The semantic relevance of each object $o_i$ in the map to the target $o_{target}$ is determined by a binary value $S(o_i, o_{target})$, indicated by a LLM. This setup ensures that all objects deemed semantically related to the target are assigned a non-zero attention weight \zdx{and remain unaffected by the object affinities from training data}. By employing this uniform attention mechanism, the module guarantees that the agent considers all potentially relevant objects, thus improving its adaptability and performance in unfamiliar settings.


The generalized attention weight $A_{G_i}$ for each object $o_i$ is calculated as:
\begin{equation}
A_{G_i} = \frac{S(o_i, o_{target})}{\sum_{j=1}^{M}S(o_j, o_{target})}.
\label{eq:3}
\end{equation}
In contrast to Equ.~\eqref{eq:2}, this formulation operates on discrete binary relevance judgments $S(o_i, o_{target}) \in \{0,1\}$ from LLMs, where the function $S$ indicates semantic relevance without considering similarity magnitudes. While Equ.~\eqref{eq:2} learns continuous affinity patterns from training data through learnable parameters $\mathbf{W}_q$ and $\mathbf{W}_k$, Equ.~\eqref{eq:3} leverages fixed semantic knowledge from pre-trained language models, ensuring uniform attention distribution among all semantically relevant objects. The final output of the generalized affinities module is a normalized vector, which directs the agent's focus towards objects of interest specified by LLM for better generalization.

\subsection{Dynamic Fusion Module}
The dynamic fusion module synergizes the outputs from the generalized affinities module ($A_{G}$) and the experiential affinities module ($A_{E}$), finely tuning the balance between learned patterns and semantic guidance for optimal navigation performance. Specifically, it adjusts the contributions of $A_{G_i}$ and $A_{E_i}$, the attention scores for an object $o_i$ from the respective modules, ensuring an adaptive final attention mechanism.

This adaptation is driven by the guidance ratio $\gamma$, which is dynamically modulated based on the temporal context $\mathcal{H}$ and, when available, additional environmental factors $\mathcal{E}$.

In architectures utilizing RNNs to encode the current state, $\mathcal{H}$ includes the hidden states of the RNNs. For non-RNN architectures, $\mathcal{H}$ encompasses the agent's past trajectory and relevant environmental data, such as explored regions and observed layouts. $\mathcal{E}$ further enriches this context with external cues, aiding in the precise adjustment of $\gamma$.

\begin{figure}
  \centering
  \includegraphics[width=\columnwidth]{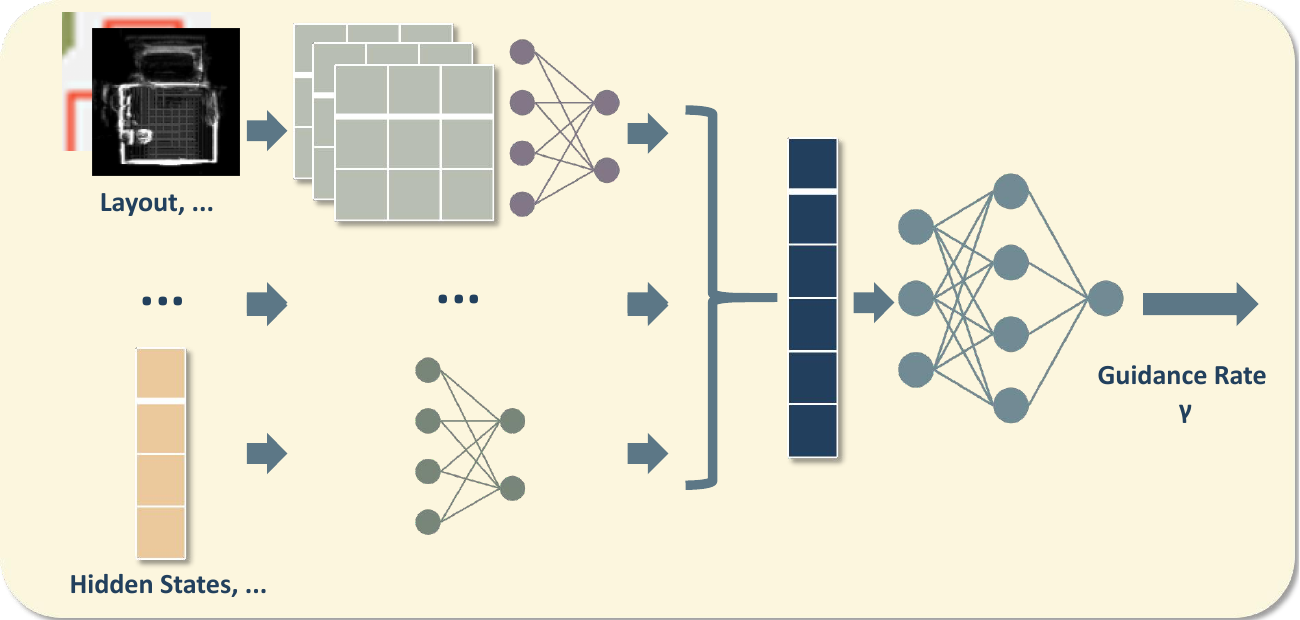} 
  \caption{\textbf{The Architecture of Dynamic Fusion Module.} The input of the dynamic fusion module could be several environmental and temporal contexts, the encoders of which are dependent on their modalities. All of the flattened features will be further concatenated together before undergoing the final MLP and outputting the guidance rate.} 
  \label{fig:dynamic}
  \vspace{-0.5cm}
\end{figure}

\re{Add some details about the rnn network.}{
The key idea involves extracting features that capture historical and environmental contexts, with processing tailored to the specific modality of each context as shown in Fig. \ref{fig:dynamic}. 
All available context features are concatenated and fed into the final MLP to obtain the dynamic rate $\gamma$.
}
\zdx{Building on this design, $\gamma$ is dynamically adjusted according to the context: it decreases in familiar environments where experiential patterns are consistent, and increases in novel or complex settings where LLMs provide more effective navigational insights. This dynamic fusion of affinity scores enables the system to adaptively employ the most effective navigation strategy for the given scenario:}

\begin{equation}
A_{F_i} = \gamma \cdot A_{G_i} + (1 - \gamma) \cdot A_{E_i}.
\label{eq:affinity}
\end{equation}

Through this mechanism, the dynamic fusion module ensures that the final affinity scores ($A_{F_i}$) for each object $o_i$ are flexibly adapted to the navigation task's specific needs. By leveraging both experiential patterns and semantic guidance, the module enhances the agent's ability to efficiently navigate across diverse environmental contexts, capitalizing on the strengths of both submodules.

\subsection{Integration with Downstream Policy}
The LOAT framework's affinity scores are systematically integrated into downstream policies to intensify the focus on objects pertinent to the navigation goal. This integration applies slightly different strategies for map-based and graph-based policies, optimizing the agent's navigational efficacy.

\textbf{Map-based Policy:} Let $S_m \in \mathbb{R}^{H 	imes W \times C}$ represent a multi-channel semantic map, where $H$ and $W$ are the spatial dimensions and $C$ is the number of object categories. Each channel $c \in \{1, 2, ..., C\}$ corresponds to a specific object category (e.g., "chair", "table", "door") and contains a 2D grid where each cell value indicates the presence or confidence of that object category at the corresponding spatial location. This representation is a classical grid-based semantic map enhanced with categorical channels, not a neural feature map.

The affinity score for each object category $c$, derived from the dynamic fusion in Equ.~\eqref{eq:affinity}, is represented by $A_{F_c} \in \mathbb{R}$. The activation of channel $c$ in response to the affinity is computed as:
\begin{equation}
    Activation(c) = A_{F_c} \cdot S_m[:,:,c],
\end{equation}
where $S_m[:,:,c] \in \mathbb{R}^{H 	imes W}$ denotes the entire 2D spatial channel for object category $c$, and the multiplication is element-wise across all spatial locations. This operation scales the entire channel by the corresponding affinity score, creating a search bias that emphasizes spatial regions containing objects deemed relevant to the target.

\textbf{Graph-based Policy:} 
For a topological graph $G$, with node $N$ representing a distinct area or a set of objects, let $A_{F_N}$ denote the averaged affinity scores for objects within node $N$, which is given by:
\begin{equation}
    A_{F_N} = \frac{1}{|O_N|} \sum_{o_i \in O_N} A_{Fo_i},
\end{equation}
where $O_N$ is the object set within node $N$ and $A_{Fo_i}$ is the affinity score for object $o_i$.

Let $E_N$ denote the embedding for node $N$ generated by the downstream graph-based policy. In typical graph-based navigation policies, the node embedding $E_N$ is constructed by aggregating the pre-trained text embeddings of all objects within the node:
\begin{equation}
    \begin{aligned}
    Activation(N) & =  E_N \cdot A_{F_N} \\
           & = \frac{E_N}{|O_N|} \sum_{o_i \in O_N} A_{Fo_i}.
    \end{aligned}
    \label{eq:node_activate}
\end{equation}
 
This node-wise activation process highlights nodes with high relevance to the target, guiding the policy to prioritize exploration of these strategically important nodes.

\textbf{Training LOAT with Downstream Policy:} LOAT functions as a flexible plugin compatible with diverse downstream policies without modifying their original training loss functions. LOAT operates as a preprocessing module that enhances input features (semantic maps or node embeddings) rather than altering learning objectives, allowing downstream policies to continue optimizing their original loss using enhanced inputs. Downstream policies may employ imitation learning, reinforcement learning, or other methods. When integrated with LOAT, training becomes two-stage. First, the experimental affinity module is co-trained with the downstream policy. Second, the generalized affinity module and dynamic fusion layer are introduced while freezing weights of both the experimental affinity module and downstream policy, updating only the dynamic fusion layer parameters.


By employing these strategies, LOAT significantly refines decision-making process, enhancing adaptability and generalization across diverse environments. Through targeted activation within downstream policies, the approach leverages learned environmental patterns and generalized semantic insights from LLMs, creating a robust framework for focused navigation.

\section{Experiments}
\subsection{Setup}

\subsubsection{Benchmarks and Metrics}
We evaluate the proposed method across Habitat~\cite{habitat19iccv} and AI2-THOR~\cite{ai2thor} simulators, including three benchmarks emphasizing object-goal navigation: ALFRED~\cite{shridhar2020alfred}, Habitat ObjectNav~\cite{habitatchallenge2022}, and SAVN-NAV~\footnote{A benchmark in AI2-THOR collected by Wortsman et. al \cite{wortsman2019savn}, referred as SAVN-NAV}. 
To evaluate the methods, we use several key metrics:
\textbf{SR (Success Rate)}: Percentage of episodes where the agent successfully completes the task by reaching the target object.
\textbf{SPL (Success weighted by Path Length)}: Ratio of the shortest path to the actual path taken, adjusted for success to assess navigation efficiency.
\textbf{PLWSR (Path Length Weighted Success Rate)}: Evaluates success while penalizing longer paths, enhancing efficiency assessment.
\textbf{GC (Goal Completion)}: Indicates whether the agent achieved the task objective.
\textbf{PLWGC (Path Length Weighted Goal Completion)}: Weights goal completion by the optimality of the path taken.
\textbf{GFR (Goal Found Rate)}: Measures the agent's ability to locate the target object based on high-level instructions, indicating navigation success apart from completion.

\begin{table*}[t]
    \caption{Comparison with the state-of-the-art Methods in ALFRED.}
    \label{tab:alfred_res}
    \centering
    \begin{tabular}{lcccccccc}
    \toprule
    \textbf{Method} & \multicolumn{4}{c}{\textbf{Tests Seen}} & \multicolumn{4}{c}{\textbf{Tests Unseen}} \\
    \cmidrule{2-9}
     & \textbf{SR} & \textbf{PLWSR} & \textbf{GC} & \textbf{PLWGC} & \textbf{SR} & \textbf{PLWSR} & \textbf{GC} & \textbf{PLWGC} \\
    \midrule
    HLSM\cite{blukis2022hlsm} & 29.94 & 8.74 & 41.21 & 14.58 & 20.27 & 5.55 & 30.31 & 9.99 \\
    FILM\cite{min2021film} & 27.67 & 11.23 & 38.51 & 15.06 & 24.46 & 10.55 & 36.37 & 14.30 \\
    LGS-RPA\cite{murray2022lgs-rpa} & 33.01 & 16.65 & 41.71 & 24.49 & 27.80 & 12.92 & 38.55 & 20.01 \\
    EPA\cite{liu2022epa} & 39.96 & 2.56 & 44.14 & 3.47 & 36.07 & 2.92 & 39.54 & 3.91 \\
    Prompter\cite{inoue2022prompter} & 49.38 & 23.47 & 55.90 & 29.06 & 42.64 & 19.49 & 59.55 & 25.00 \\
    CAPEAM\cite{kim2023capeam} & 47.36 & 19.03 & 54.38 & 23.78 & 43.69 & 17.64 & 54.66 & 22.76 \\
    LLM-Planner\cite{song2023llmplanner} & 18.20 & - & 26.77 & - & 16.42 & - & 23.37 & - \\
    LOAT-P (ours) & \textbf{56.03} & \textbf{28.59} & \textbf{65.36} & \textbf{33.54} & \textbf{54.22} &  \textbf{28.12} & \textbf{63.85} & \textbf{33.51} \\
    \bottomrule
    \end{tabular}
    \vspace{-0.5cm}
\end{table*}

\subsubsection{Implementation Details and Baselines}
We utilize \textit{paraphrase-MiniLM-L6-v2} \cite{reimers2019sentencebert} to compute text embeddings for category names, producing 384-dimensional dense vectors. To assess object affinities, GPT-4 is introduced to store semantic scores for faster navigation. Two 2-layer MLPs with ReLU are employed to calculate the guidance ratio $\gamma$.


For ALFRED tasks, we enhance the FILM\cite{min2021film} system with LOAT for semantic map activation. We replace Mask R-CNN with a fine-tuned DINO\cite{caron2021dino} model for object segmentation because DINO provides more robust object detection in diverse indoor environments, which is essential for accurate semantic map construction that LOAT relies upon. Additionally, we adapt a Prompter-inspired\cite{inoue2022prompter} exploration strategy as it demonstrates superior performance in instruction-following tasks, ensuring fair comparison by using strong baseline components. These modifications are applied consistently across all compared methods to maintain experimental fairness. This enhanced system, called LOAT-P, offers an integrated solution for improved navigation and object interaction in AI2-THOR environments.

LOAT-P is compared to several established methods, including HLSM~\cite{blukis2022hlsm}, which uses a spatial-semantic voxel map for environment modeling; FILM~\cite{min2021film}, known for its spatial memory and semantic search capabilities; LGS-RPA~\cite{murray2022lgs-rpa}, which employs landmark-guided search; EPA~\cite{liu2022epa}, featuring neural-symbolic planning; Prompter \cite{inoue2022prompter}, based on template-driven planning; CAPEAM \cite{kim2023capeam}, which integrates spatial and state-change information; and LLM-Planner \cite{song2023llmplanner}, a training-free approach leveraging large language models for few-shot planning.

We also evaluate the LOAT framework with a graph-based policy from the HOZ\cite{zhang2021hierarchical} system on the SAVN-NAV navigation tasks\cite{wortsman2019savn}. The activated hierarchical graphs are then fed into an A3C\cite{mnih2016a3c} policy and trained with reinforcement learning. We compare the LOAT-enhanced HOZ with a naive A3C policy that utilizes a simple visual embedding layer, as well as other graph-centric methods like ORG\cite{du2020org} and SP\cite{yang2018sp}, which establish object semantic relationships.

For Habitat ObjectNav tasks, we adapt a PEANUT-inspired model\cite{zhai2023peanut}, trained with the LOAT framework. For evaluating the LOAT-enhanced PEANUT, we benchmark it against  PEANUT and DD-PPO\cite{wijmans2020ddppo} trained on 20,000 human demonstrations.

\subsection{Results}

ALFRED experiment results are shown in Table \ref{tab:alfred_res}. LOAT-P achieves SOTA performance across all metrics, with a notable 10\% increase in SR. While some improvements appear modest in absolute terms, they represent meaningful advances in object navigation where even small gains often require substantial algorithmic innovations. Notably, the SR discrepancy between familiar and unfamiliar environments is markedly reduced, showcasing LOAT's efficacy in applying generalized object affinities from LLMs in novel settings.


Following~\cite{zhang2021hierarchical}, we employ SAVN-NAV navigation tasks~\cite{wortsman2019savn} with random selection of agent's initial position and goal item, using five trials per scenario. Table \ref{tab:robothor_res} displays results for all targets and a subset with optimal path lengths above five steps. Integrating LOAT into the HOZ system improved both SPL and SR metrics, surpassing previously reproduced results and underscoring LOAT's effectiveness within graph-based policy.

\begin{table}
\caption{Comparison in SAVN-NAV navigation tasks in AI2-THOR.}
\label{tab:robothor_res}
\centering
\begin{tabular}{lcccc}
\toprule
\textbf{Method} & \multicolumn{2}{c}{\textbf{All}} & \multicolumn{2}{c}{\textbf{Long Trials (L $\geq$ 5)}} \\
\cmidrule{2-5} 
                & \textbf{SPL}  & \textbf{SR}   & \textbf{SPL}  & \textbf{SR}             \\
\midrule
Random          & 1.73         & 3.56                & 0.07         & 0.27                \\
A3C \cite{mnih2016a3c}(baseline)  & 33.78        & 57.35               & 30.65        & 45.77               \\
SP \cite{yang2018sp}         & 37.01        & 62.16               & 34.17        & 50.86               \\
ORG \cite{du2020org}        & 38.42        & 66.38               & 36.26        & 55.55               \\
HOZ \cite{zhang2021hierarchical} & 38.80      & 72.20               & 38.83        & 64.05               \\
HOZ w/ LOAT      & \textbf{39.56}       & \textbf{73.12}    & \textbf{39.68} & \textbf{65.26}    \\ 
\bottomrule
\end{tabular}
\end{table}

\begin{table}
    \centering
    \caption{Results in Habitat ObjectNav in Val split.}
    \label{tab:habitat_res}
    \begin{tabular}{lcc}
    \toprule
    \textbf{Method} & \textbf{SPL} & \textbf{SR} \\
    \midrule
    DD-PPO \cite{wijmans2020ddppo} & 0.20 & 0.52  \\
    Habitat-Web \cite{ramrakhya2022habitatweb} & 0.22 & 0.55 \\
    ProcTHOR \cite{deitke2022procthor} & 0.32 & 0.54 \\
    PIRLNav\cite{ramrakhya2023pirlnav} & 0.28 & 0.62\\
    PEANUT \cite{zhai2023peanut} & 0.30 & 0.55 \\
    PEANUT w/ LOAT  & \textbf{0.32} & \textbf{0.63} \\
    \bottomrule
\end{tabular}
\end{table}


Habitat ObjectNav validation results in Table \ref{tab:habitat_res} demonstrate significant performance enhancement from incorporating LOAT. The PEANUT w/ LOAT method exhibits notable improvements in both SPL and SR, suggesting that LOAT not only optimizes policy learning but also significantly boosts navigation capabilities in complex environments.


Comprehensive analysis across three benchmark evaluations reveals that LOAT excels in three key aspects:

\begin{enumerate}
    \item Improves effectiveness and generalization of navigation systems. LOAT combines experienced object affinities with LLM semantic understanding, leading to higher success rates, reduced navigation pathways, and improved generalization to new environments.
    \item Compatible with current SOTA methods, LOAT enhances performance in various navigation contexts by working well with both graph-based and map-based policies, improving performance regardless of the downstream policy mechanism.  
    \item Offers targeted guidance for navigating towards small, hard-to-locate or unseen objects. By leveraging object affinities from both training experiences and LLM insights, it efficiently detects objects difficult to identify without context from relevant easier-to-find objects. For example, LOAT locates a \textit{"fork"} by first identifying a nearby \textit{"dining table"} or \textit{"kitchen counter"} where forks commonly appear.
\end{enumerate}

\subsection{Ablation Studies}
\subsubsection{Impact of the LOAT Framework in ALFRED\cite{shridhar2020alfred}}

The LOAT-P system dramatically increased ALFRED's performance in previous experiments. To differentiate the impact of the LOAT framework from other improvements, such as advanced segmentation methods, LOAT is rigorously evaluated by integrating it into the Prompter \cite{inoue2022prompter} and FILM \cite{min2021film} on the ALFRED benchmark, respectively. Table \ref{tab:ablation_loat_overall} shows enhancements across all measured metrics upon incorporating the LOAT framework. Notably, the improvement in GFR is a testament to LOAT's ability to significantly enhance the agent's proficiency in identifying target objects. This underscores the framework's effectiveness in augmenting navigational and object-identification capabilities with insights from both LLM-derived object affinities and training-time preferences, thereby contributing to the overall performance uplift.


\begin{table}
    \caption{Performance with/without LOAT Integration in ALFRED. 
    }
    \label{tab:ablation_loat_overall}
    \centering
    \begin{tabular}{lccccccccc}
    \toprule
    \textbf{Method} & \multicolumn{2}{c}{\textbf{Valid Seen}} & \multicolumn{2}{c}{\textbf{Valid Unseen}} \\
    \cmidrule{2-5} 
     & \textbf{SR} & \textbf{GFR} & \textbf{SR} & \textbf{GFR} \\
    \midrule
    FILM\cite{min2021film} & 20.10 & 56.15 & 23.66 & 58.41 \\
    FILM\cite{min2021film} w/ LOAT & 22.78  & 56.27  & 26.46  & 60.37  \\
    Prompter\cite{inoue2022prompter} & 51.04 & 65.04 & 52.20 & 73.66 \\
    Prompter\cite{inoue2022prompter} w/ LOAT & 53.47  & 66.63  & 53.78  & 75.85  \\
    \bottomrule
    \end{tabular}
\end{table}


\subsubsection{Impact of LOAT Submodules}


The LOAT framework's integration of Generalized Affinities (G.A.) and Experiential Affinities (E.A.) modules enhances navigation and task execution, as shown in Table \ref{tab:ablation_loat_submodule}. The G.A. module slightly improves success rates through general knowledge guidance, but shows limitations in complex scenarios without learned experience. The E.A. module improves outcomes in both short and long trials by applying specific past experiences.


Integration of both modules yields the highest performance improvements. While G.A. provides a broad knowledge base, E.A. extracts targeted knowledge from training, optimizing both general principles and specific experiences. Beyond this parallel integration of LLM commonsense and experiential object affinities, we also explore using LLM commonsense as a constraint. This comprehensive approach enables LOAT to navigate complex environments more effectively, demonstrating the importance of integrating diverse knowledge sources for advanced decision-making and problem-solving capabilities.

\begin{table}[tb]
    \caption{Comparative evaluation of HOZ incorporating LOAT framework variants, Experiential Affinities (E.A.) and Generalized Affinities (G.A.) modules.
    }
    \label{tab:ablation_loat_submodule}
    \centering
    \resizebox{\columnwidth}{!}{%
    \begin{tabular}{lcccc}
    \toprule
    \textbf{Method} & \multicolumn{2}{c}{\textbf{All Trials}} & \multicolumn{2}{c}{\textbf{Long Trials (L $\geq$ 5)}} \\ \cmidrule{2-5} 
                    & \textbf{SPL (\%)} & \textbf{Success (\%)} & \textbf{SPL (\%)} & \textbf{Success (\%)} \\ 
                    \midrule
    Baseline (HOZ)\cite{wortsman2019savn} & 38.80 & 72.20 & 38.83 & 64.05 \\
    w/ G.A. Module & 38.94  & 72.40  & 38.78  & 64.05  \\
    w/ E.A. Module & 39.03  & 72.51  & 39.08  & 64.35  \\ 
    w/ Full LOAT & \textbf{39.56} & \textbf{73.12} & \textbf{39.68} & \textbf{65.26} \\ 
    \bottomrule
    \end{tabular}%
    }
    \vspace{-0.5cm}
\end{table}

\begin{figure}[htbp]
  \centering
  \begin{subfloat}[\footnotesize LOAT-P w/o G.A.]{%
    \includegraphics[height=3.67cm, keepaspectratio, page=1]{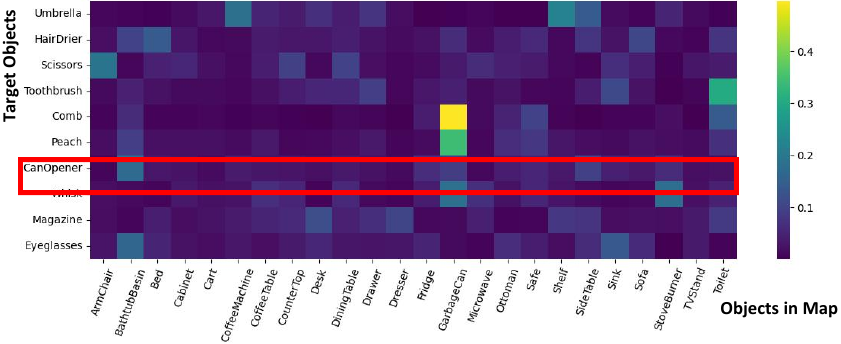}
    \label{fig:dist_extra_page1}
    }
  \end{subfloat}
  \begin{subfloat}[\footnotesize LOAT-P]{%
    \includegraphics[height=3.67cm, keepaspectratio, page=2]{images/attn_images/distribution_extra_crop.pdf}
    \label{fig:dist_extra_page2}
    }
  \end{subfloat}
  \caption{\textbf{Predicted Distribution for Out-of-domain Objects.} In (a), models without the generalized affinities module may make less acceptable assumptions, such as linking a can opener with a bathtub basin. Using LLM-derived knowledge, models in (b) make more accurate predictions, such as positioning can openers near cabinets or coffee tables.}
  \label{fig:dist_extra}
\end{figure}
\vspace{-1em}

\subsection{Evaluation on Out-of-Domain Objects}

Since LOAT employs text embeddings instead of one-hot encodings for task representation, we examine model performance on unseen targets without training. We assessed out-of-domain object recognition using 300 pre-collected semantic maps from AI2-THOR environments, predicting potential locations for unseen targets based on category names: \textit{"Umbrella", "HairDrier", "Scissors", "Toothbrush", "Comb", "Peach", "CanOpener", "Whisk", "Magazine", "Eyeglasses"}. The model identifies nearest objects in the map to these targets, applying a distance threshold of one-fifteenth of the map's total resolution. We focused on assessing semantic logic behind probability distributions to determine if predicted locations were plausible within domestic contexts. Fig. \ref{fig:dist_extra} shows that models relying solely on experience transfer may predict uniform distributions, resulting in semantically incongruous predictions (\eg associating a can opener with a bathtub basin). LOAT-integrated models better incorporate commonsense reasoning from LLMs, producing logically coherent probability distributions (\eg placing a can opener near cabinets or coffee tables) reflecting intuitive home placement expectations.

\subsection{Navigation in Real Environments}

\begin{figure}[t]
    \centering
    \includegraphics[width=\linewidth]{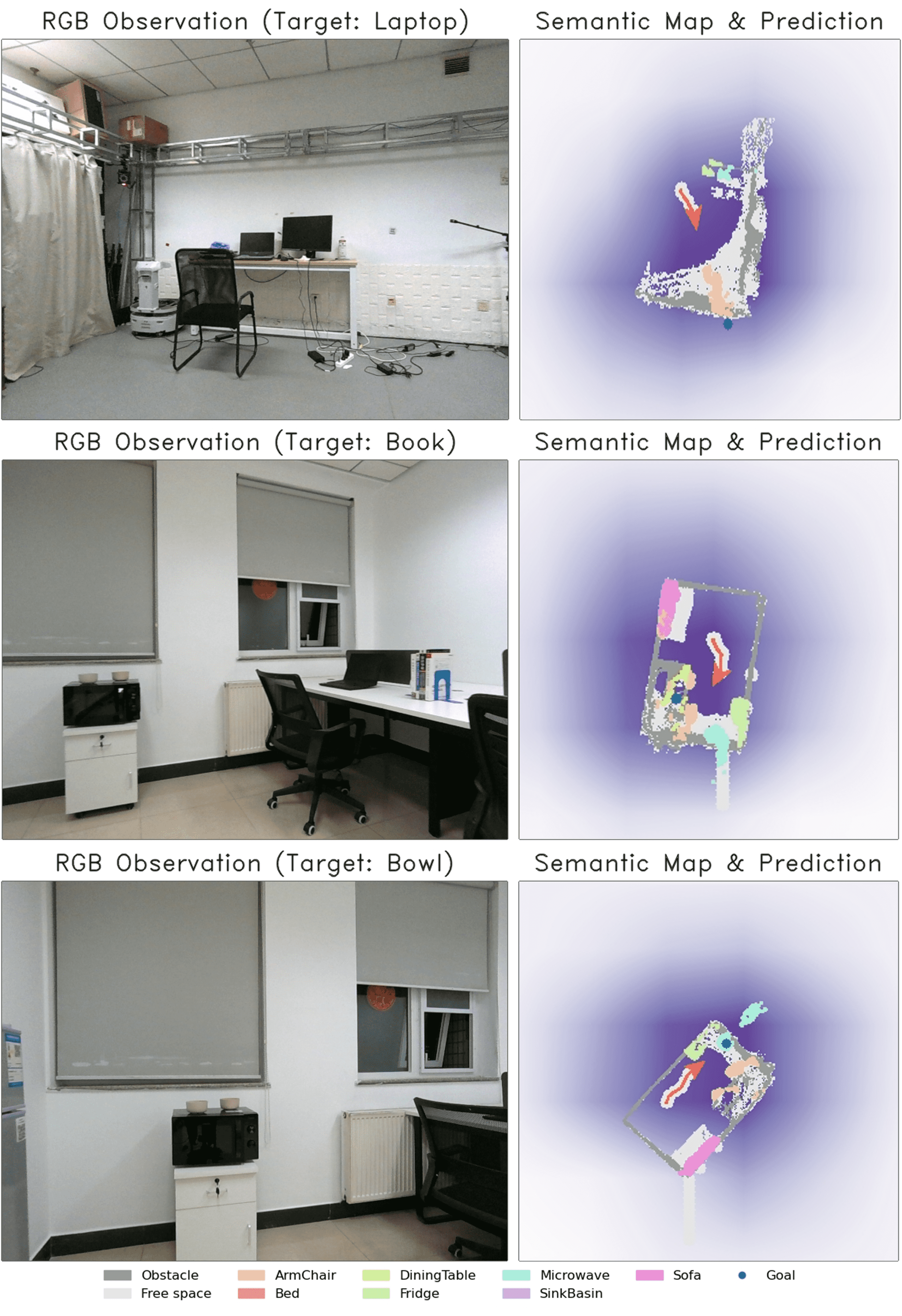}
    \caption{\textbf{Real-world Navigation Examples.} In real world experiments, model with LOAT is able to make reasonable predictions without extra training.}
    \label{fig:real_env}
    \vspace{-0.5cm}
\end{figure}


We conduct real-world navigation tasks with the LoCoBot wx250s platform, powered by an Intel NUC Mini PC running ROS. The robot connects to an NVIDIA GeForce RTX 3060 laptop hosting the LOAT-P navigation system, while an Intel RealSense D455 camera captures RGB and depth images for environmental perception. The LOAT-P system was deployed in two laboratory-adapted scenes simulating household environments with distinct semantic divisions. It located five object categories—	\textit{Apple, Book, Bowl, Cup, and Laptop}—across three trials per scene from various initial positions. The system transitioned directly from simulation (trained on ALFRED dataset) to real-world applications without further training.


Results indicated successful location of all target objects, demonstrating LOAT's effective generalization to unseen environments. Visual examples, such as accurate bowl identification near the microwave, are shown in Fig. \ref{fig:real_env}. These findings confirm the robustness and adaptability of LOAT-P navigation system in real-world scenarios.

\section{Conclusions}

In this paper, we propose the LOAT framework, a novel approach that integrates LLM-derived object semantics with historical experiential object affinities to enhance robotic navigation. LOAT significantly improves navigation in environments like AI2-THOR and Habitat across three benchmarks, demonstrating excellent navigation performance and generalization ability in unseen scenarios and targets. In addition, as a flexible plugin, it can be combined with different types of downstream policy, such as metric-map-based and topological-graph-based policies. 

While LOAT’s performance can be influenced by the diversity of object categories in the underlying semantic map, extending LOAT’s dual-module strategy to open vocabulary navigation is still challenging. Such an approach holds promise for expanding its adaptability to a broader range of environments and tasks, further advancing autonomous navigation capabilities.


\bibliographystyle{IEEEtran}
\bibliography{main}

\begin{thebibliography}{10}
\providecommand{\url}[1]{#1}
\csname url@samestyle\endcsname
\providecommand{\newblock}{\relax}
\providecommand{\bibinfo}[2]{#2}
\providecommand{\BIBentrySTDinterwordspacing}{\spaceskip=0pt\relax}
\providecommand{\BIBentryALTinterwordstretchfactor}{4}
\providecommand{\BIBentryALTinterwordspacing}{\spaceskip=\fontdimen2\font plus
\BIBentryALTinterwordstretchfactor\fontdimen3\font minus
  \fontdimen4\font\relax}
\providecommand{\BIBforeignlanguage}[2]{{%
\expandafter\ifx\csname l@#1\endcsname\relax
\typeout{** WARNING: IEEEtran.bst: No hyphenation pattern has been}%
\typeout{** loaded for the language `#1'. Using the pattern for}%
\typeout{** the default language instead.}%
\else
\language=\csname l@#1\endcsname
\fi
#2}}
\providecommand{\BIBdecl}{\relax}
\BIBdecl

\bibitem{1sun2024survey}
J.~Sun, J.~Wu, Z.~Ji, and Y.-K. Lai, ``A survey of object goal navigation,''
  \emph{IEEE Transactions on Automation Science and Engineering}, vol.~22, pp.
  2292--2308, 2025.

\bibitem{chen2023not-train}
J.~Chen, G.~Li, S.~Kumar, B.~Ghanem, and F.~Yu, ``How to not train your dragon:
  Training-free embodied object goal navigation with semantic frontiers,'' in
  \emph{Robotics: Science and Systems XIX, Daegu, Republic of Korea, July
  10-14, 2023}, K.~E. Bekris, K.~Hauser, S.~L. Herbert, and J.~Yu, Eds., 2023.

\bibitem{zhang2021hierarchical}
S.~Zhang, X.~Song, Y.~Bai, W.~Li, Y.~Chu, and S.~Jiang, ``Hierarchical
  object-to-zone graph for object navigation,'' in \emph{Proceedings of the
  IEEE/CVF International Conference on Computer Vision}, 2021, pp.
  15\,130--15\,140.

\bibitem{zhou2023esc}
K.~Zhou, K.~Zheng, C.~Pryor, Y.~Shen, H.~Jin, L.~Getoor, and X.~E. Wang, ``Esc:
  Exploration with soft commonsense constraints for zero-shot object
  navigation,'' in \emph{International Conference on Machine Learning}.\hskip
  1em plus 0.5em minus 0.4em\relax PMLR, 2023, pp. 42\,829--42\,842.

\bibitem{rajvanshi2023saynav}
A.~Rajvanshi, K.~Sikka, X.~Lin, B.~Lee, H.-P. Chiu, and A.~Velasquez, ``Saynav:
  Grounding large language models for dynamic planning to navigation in new
  environments,'' in \emph{Proceedings of the International Conference on
  Automated Planning and Scheduling}, vol.~34, 2024, pp. 464--474.

\bibitem{zhou2023navgpt}
G.~Zhou, Y.~Hong, and Q.~Wu, ``Navgpt: Explicit reasoning in
  vision-and-language navigation with large language models,'' in
  \emph{Proceedings of the AAAI Conference on Artificial Intelligence},
  vol.~38, no.~7, 2024, pp. 7641--7649.

\bibitem{mat}
B.~Li, H.~Li, Y.~Zhu, and D.~Zhao, ``Mat: Morphological adaptive transformer
  for universal morphology policy learning,'' \emph{IEEE Transactions on
  Cognitive and Developmental Systems}, vol.~16, no.~4, pp. 1611--1621, 2024.

\bibitem{Haoran2020}
H.~Li, Q.~Zhang, and D.~Zhao, ``Deep reinforcement learning-based automatic
  exploration for navigation in unknown environment,'' \emph{IEEE Transactions
  on Neural Networks and Learning Systems}, vol.~31, no.~6, pp. 2064--2076,
  2020.

\bibitem{chen2021decisiontransformer}
L.~Chen, K.~Lu, A.~Rajeswaran, K.~Lee, A.~Grover, M.~Laskin, P.~Abbeel,
  A.~Srinivas, and I.~Mordatch, ``Decision transformer: Reinforcement learning
  via sequence modeling,'' in \emph{Advances in Neural Information Processing
  Systems}, vol.~34.\hskip 1em plus 0.5em minus 0.4em\relax Curran Associates,
  Inc., 2021, pp. 15\,084--15\,097.

\bibitem{lin2021multimodal}
C.~Lin, Y.~Jiang, J.~Cai, L.~Qu, G.~Haffari, and Z.~Yuan, ``Multimodal
  transformer with variable-length memory for vision-and-language navigation,''
  in \emph{Computer Vision -- ECCV 2022}, 2022, pp. 380--397.

\bibitem{zeng2024poliformer}
K.-H. Zeng, Z.~Zhang, K.~Ehsani, R.~Hendrix, J.~Salvador, A.~Herrasti,
  R.~Girshick, A.~Kembhavi, and L.~Weihs, ``Poliformer: Scaling on-policy rl
  with transformers results in masterful navigators,'' in \emph{Proceedings of
  the Conference on Robot Learning (CoRL)}, 2024.

\bibitem{chen2022duet}
S.~Chen, P.-L. Guhur, M.~Tapaswi, C.~Schmid, and I.~Laptev, ``Think global, act
  local: Dual-scale graph transformer for vision-and-language navigation,'' in
  \emph{Proceedings of the IEEE/CVF Conference on Computer Vision and Pattern
  Recognition}, 2022, pp. 16\,537--16\,547.

\bibitem{min2021film}
S.~Y. Min, D.~S. Chaplot, P.~K. Ravikumar, Y.~Bisk, and R.~Salakhutdinov,
  ``{FILM:} following instructions in language with modular methods,'' in
  \emph{International Conference on Learning Representations}, 2022.

\bibitem{zhai2023peanut}
A.~J. Zhai and S.~Wang, ``Peanut: predicting and navigating to unseen
  targets,'' in \emph{Proceedings of the IEEE/CVF International Conference on
  Computer Vision}, 2023, pp. 10\,926--10\,935.

\bibitem{chen2024common}
Y.~Chen, X.~Zhang, Y.~Chen, D.~Zhao, Y.~Zhao, Z.~Zhao, and P.~Hu, ``Common
  sense language-guided exploration and hierarchical dense perception for
  instruction following embodied agents,'' in \emph{2024 IEEE International
  Conference on Multimedia and Expo (ICME)}, 2024, pp. 1--6.

\bibitem{Yamauchi_2002}
B.~Yamauchi, ``A frontier-based approach for autonomous exploration,'' in
  \emph{Proceedings 1997 IEEE International Symposium on Computational
  Intelligence in Robotics and Automation CIRA'97. 'Towards New Computational
  Principles for Robotics and Automation'}, 1997, pp. 146--151.

\bibitem{gadre2022cow}
S.~Y. Gadre, M.~Wortsman, G.~Ilharco, L.~Schmidt, and S.~Song, ``Cows on
  pasture: Baselines and benchmarks for language-driven zero-shot object
  navigation,'' in \emph{2023 IEEE/CVF Conference on Computer Vision and
  Pattern Recognition (CVPR)}, 2023, pp. 23\,171--23\,181.

\bibitem{Ramakrishnan22}
S.~K. Ramakrishnan, D.~S. Chaplot, Z.~Al-Halah, J.~Malik, and K.~Grauman,
  ``Poni: Potential functions for objectgoal navigation with interaction-free
  learning,'' in \emph{2022 IEEE/CVF Conference on Computer Vision and Pattern
  Recognition (CVPR)}, 2022, pp. 18\,868--18\,878.

\bibitem{topiwala2018frontier}
A.~Topiwala, P.~Inani, and A.~Kathpal, ``Frontier based exploration for
  autonomous robot,'' \emph{CoRR}, vol. abs/1806.03581, 2018.

\bibitem{yuanwen2025}
Y.~Chen, H.~Li, Y.~Chen, and D.~Zhao, ``Leaffordnav: Enhancing open-vocabulary
  mobile manipulation with llm-guided exploration and affordance-aware
  navigation,'' in \emph{2025 IEEE International Conference on Multimedia and
  Expo (ICME)}, 2025.

\bibitem{Yu_l3mvn2023}
B.~Yu, H.~Kasaei, and M.~Cao, ``L3mvn: Leveraging large language models for
  visual target navigation,'' in \emph{2023 IEEE/RSJ International Conference
  on Intelligent Robots and Systems (IROS)}.\hskip 1em plus 0.5em minus
  0.4em\relax IEEE, 2023.

\bibitem{inoue2022prompter}
Y.~Inoue and H.~Ohashi, ``Prompter: Utilizing large language model prompting
  for a data efficient embodied instruction following,'' \emph{CoRR}, vol.
  abs/2211.03267, 2022.

\bibitem{robogpt}
Y.~Chen, W.~Cui, Y.~Chen, M.~Tan, X.~Zhang, J.~Liu, H.~Li, D.~Zhao, and
  H.~Wang, ``Robogpt: an llm-based long-term decision-making embodied agent for
  instruction following tasks,'' \emph{IEEE Transactions on Cognitive and
  Developmental Systems}, pp. 1--11, 2025.

\bibitem{song2023llmplanner}
C.~H. Song, J.~Wu, C.~Washington, B.~M. Sadler, W.-L. Chao, and Y.~Su,
  ``Llm-planner: Few-shot grounded planning for embodied agents with large
  language models,'' in \emph{Proceedings of the IEEE/CVF International
  Conference on Computer Vision}, 2023, pp. 2998--3009.

\bibitem{habitat19iccv}
M.~Savva, A.~Kadian, O.~Maksymets, Y.~Zhao, E.~Wijmans, B.~Jain, J.~Straub,
  J.~Liu, V.~Koltun, J.~Malik, D.~Parikh, and D.~Batra, ``Habitat: {A}
  {P}latform for {E}mbodied {AI} {R}esearch,'' in \emph{Proceedings of the
  IEEE/CVF International Conference on Computer Vision (ICCV)}, 2019.

\bibitem{ai2thor}
E.~Kolve, R.~Mottaghi, W.~Han, E.~VanderBilt, L.~Weihs, A.~Herrasti, D.~Gordon,
  Y.~Zhu, A.~Gupta, and A.~Farhadi, ``{AI2-THOR: An Interactive 3D Environment
  for Visual AI},'' \emph{arXiv}, 2017.

\bibitem{shridhar2020alfred}
M.~Shridhar, J.~Thomason, D.~Gordon, Y.~Bisk, W.~Han, R.~Mottaghi,
  L.~Zettlemoyer, and D.~Fox, ``Alfred: A benchmark for interpreting grounded
  instructions for everyday tasks,'' in \emph{Proceedings of the IEEE/CVF
  Conference on Computer Vision and Pattern Recognition}, 2020, pp.
  10\,740--10\,749.

\bibitem{habitatchallenge2022}
K.~Yadav, S.~K. Ramakrishnan, A.~Gokaslan, O.~Maksymets, R.~Jain, R.~Ramrakhya,
  A.~X. Chang, A.~Clegg, M.~Savva, E.~Undersander, D.~S. Chaplot, and D.~Batra,
  ``Habitat challenge 2022,'' \url{https://aihabitat.org/challenge/2022/},
  2022.

\bibitem{wortsman2019savn}
M.~Wortsman, K.~Ehsani, M.~Rastegari, A.~Farhadi, and R.~Mottaghi, ``Learning
  to learn how to learn: Self-adaptive visual navigation using meta-learning,''
  in \emph{Proceedings of the IEEE/CVF Conference on Computer Vision and
  Pattern Recognition}, 2019, pp. 6750--6759.

\bibitem{blukis2022hlsm}
V.~Blukis, C.~Paxton, D.~Fox, A.~Garg, and Y.~Artzi, ``A persistent spatial
  semantic representation for high-level natural language instruction
  execution,'' in \emph{Conference on Robot Learning}.\hskip 1em plus 0.5em
  minus 0.4em\relax PMLR, 2022, pp. 706--717.

\bibitem{murray2022lgs-rpa}
M.~Murray and M.~Cakmak, ``Following natural language instructions for
  household tasks with landmark guided search and reinforced pose adjustment,''
  \emph{IEEE Robotics and Automation Letters}, vol.~7, no.~3, pp. 6870--6877,
  2022.

\bibitem{liu2022epa}
X.~Liu, H.~Palacios, and C.~Muise, ``A planning based neural-symbolic approach
  for embodied instruction following,'' \emph{Interactions}, vol.~9, no.~8,
  p.~17, 2022.

\bibitem{kim2023capeam}
B.~Kim, J.~Kim, Y.~Kim, C.~Min, and J.~Choi, ``Context-aware planning and
  environment-aware memory for instruction following embodied agents,'' in
  \emph{Proceedings of the IEEE/CVF International Conference on Computer
  Vision}, 2023, pp. 10\,936--10\,946.

\bibitem{reimers2019sentencebert}
N.~Reimers and I.~Gurevych, ``Sentence-bert: Sentence embeddings using siamese
  bert-networks,'' in \emph{Proceedings of the 2019 Conference on Empirical
  Methods in Natural Language Processing}.\hskip 1em plus 0.5em minus
  0.4em\relax Association for Computational Linguistics, 11 2019.

\bibitem{caron2021dino}
M.~Caron, H.~Touvron, I.~Misra, H.~J{\'e}gou, J.~Mairal, P.~Bojanowski, and
  A.~Joulin, ``Emerging properties in self-supervised vision transformers,'' in
  \emph{Proceedings of the IEEE/CVF International Conference on Computer
  Vision}, 2021, pp. 9650--9660.

\bibitem{mnih2016a3c}
V.~Mnih, A.~P. Badia, M.~Mirza, A.~Graves, T.~P. Lillicrap, T.~Harley,
  D.~Silver, and K.~Kavukcuoglu, ``Asynchronous methods for deep reinforcement
  learning,'' in \emph{Proceedings of the International Conference on Machine
  Learning}, 2016, pp. 1928--1937.

\bibitem{du2020org}
H.~Du, X.~Yu, and L.~Zheng, ``Learning object relation graph and tentative
  policy for visual navigation,'' in \emph{Computer Vision -- ECCV 2020},
  A.~Vedaldi, H.~Bischof, T.~Brox, and J.-M. Frahm, Eds.\hskip 1em plus 0.5em
  minus 0.4em\relax Cham: Springer International Publishing, 2020, pp. 19--34.

\bibitem{yang2018sp}
W.~Yang, X.~Wang, A.~Farhadi, A.~Gupta, and R.~Mottaghi, ``Visual semantic
  navigation using scene priors,'' in \emph{International Conference on
  Learning Representations}, 2019.

\bibitem{wijmans2020ddppo}
E.~Wijmans, A.~Kadian, A.~Morcos, S.~Lee, I.~Essa, D.~Parikh, M.~Savva, and
  D.~Batra, ``{DD-PPO:} learning near-perfect pointgoal navigators from 2.5
  billion frames,'' in \emph{International Conference on Learning
  Representations}, 2020.

\bibitem{ramrakhya2022habitatweb}
R.~Ramrakhya, E.~Undersander, D.~Batra, and A.~Das, ``Habitat-web: Learning
  embodied object-search strategies from human demonstrations at scale,'' in
  \emph{Proceedings of the IEEE/CVF Conference on Computer Vision and Pattern
  Recognition}, 2022, pp. 5173--5183.

\bibitem{deitke2022procthor}
M.~Deitke, E.~VanderBilt, A.~Herrasti, L.~Weihs, K.~Ehsani, J.~Salvador,
  W.~Han, E.~Kolve, A.~Kembhavi, and R.~Mottaghi, ``Procthor: Large-scale
  embodied ai using procedural generation,'' \emph{Advances in Neural
  Information Processing Systems}, vol.~35, pp. 5982--5994, 2022.

\bibitem{ramrakhya2023pirlnav}
R.~Ramrakhya, D.~Batra, E.~Wijmans, and A.~Das, ``Pirlnav: Pretraining with
  imitation and rl finetuning for objectnav,'' in \emph{Proceedings of the
  IEEE/CVF Conference on Computer Vision and Pattern Recognition}, 2023, pp.
  17\,896--17\,906.

\end{thebibliography}

\end{document}